# Compressing Neural Networks using the Variational Information Bottleneck

**Bin Dai** DAIB13@MAILS.TSINGHUA.EDU.CN
*Institute for Advanced Study*
*Tsinghua University*
*Beijing, China*

**Chen Zhu** ZHUCHEN@SHANGHAITECH.EDU.CN
*ShanghaiTech University,*
*Shanghai, China*

**David Wipf** DAVIDWIP@MICROSOFT.COM
*Microsoft Research*
*Beijing, China*

## Abstract

Neural networks can be compressed to reduce memory and computational requirements, or to increase accuracy by facilitating the use of a larger base architecture. In this paper we focus on pruning individual neurons, which can simultaneously trim model size, FLOPs, and run-time memory. To improve upon the performance of existing compression algorithms we utilize the information bottleneck principle instantiated via a tractable variational bound. Minimization of this information theoretic bound reduces the redundancy between adjacent layers by aggregating useful information into a subset of neurons that can be preserved. In contrast, the activations of disposable neurons are shut off via an attractive form of sparse regularization that emerges naturally from this framework, providing tangible advantages over traditional sparsity penalties without contributing additional tuning parameters to the energy landscape. We demonstrate state-of-the-art compression rates across an array of datasets and network architectures.

## 1. Introduction

Although extremely effective across diverse application domains, it is nonetheless well-established that many popular deep neural network architectures are over-parameterized even with respect to datasets where predictive performance is high (Denil et al., 2013). Therefore, accuracy need not suffer per se, but unnecessarily large computational and memory footprints are required for practical deployment. Hence there is tremendous potential to compress trained networks while preserving the original accuracy. Alternatively, it has also been shown that even if a given network architecture is not necessarily over-parameterized, a larger network pruned down to a smaller one of equivalent size often produces higher accuracy (Lin et al., 2017). Therefore network compression can potentially boost both efficiency and accuracy in some sense.

As reviewed in Section 2, a huge number of algorithmic pipelines have been recently proposed to instantiate some form of neural network compression; however, there is as of yet no perfect solution leaving room for new developments. In this paper, we borrow the idea of the information bottleneck (Tishby et al., 2000; Tishby & Zaslavsky, 2015), which provides a convenient mechanism for penalizing an information theoretic measure of redundancy between adjacent network layers which can be subsequently harnessed for compression.

More specifically, if we interpret these layers as forming a Markov chain, then typical training involves finding weights such that the maximal information pertaining to label $\boldsymbol{y}$ propagates from an input $\boldsymbol{x}$ to the network output. However, when a so-called information bottleneck is applied, superfluous information related to $\boldsymbol{x}$ but irrelevant for predicting $\boldsymbol{y}$ can be squeezed from the model layer-by-layer. There are multiple ways such a bottleneck could be introduced, but it is most helpful to consider strategies readily amenable to practical implementation.

To this end, we penalize the inter-layer mutual information using a variational approximation that both: (i) circumvents certain intractable integrations via a friendly bound, and (ii) reduces redundancy by aggregating useless information into certain expendable neurons using a latent sparsity-promotion mechanism. These neurons can naturally be identified and pruned to both reduce model size, FLOPs, and the run-time memory footprint. In accomplishing this, our main contributions are three-fold:

1. Beginning from the information bottleneck principle and a tractable variational approximation, we design a well-motivated neural network compression energy function that requires only a single, unavoidable tuning parameter for managing the compression/accuracy trade-off. No additional hyper-parameters for describing priors or other special constraints are required.

2. We carefully analyze an emergent tendency to accumulate useful information in a sparse set of neurons, while pushing the activations of others towards zero. Moreover, we quantify how this favoritism towards sparsity and implicit network pruning holds certain advantages over more traditional sparsity penalties that have previously been applied to network compression.

3. Finally, we present empirical results across the most common compression benchmarks, demonstrating improvement over numerous state-of-the-art approaches.

## 2. Related Network Compression Work

To obtain neural networks with low computational cost and/or memory footprint, prior work has involved designing more efficient architectures (Howard et al., 2017; Dong et al., 2017; Iandola et al., 2016), quantizing network weights (Courbariaux et al., 2016, 2015; Han et al., 2015a; Mellempudi et al., 2017; Rastegari et al., 2016), using efficient tensor or matrix decompositions to compress layers (Jaderberg et al., 2014; Zhang et al., 2016; Yu et al., 2017), or pruning existing network structures.

With respect to pruning, one option is to pre-train a network and then subsequently remove connections with small absolute values (Han et al., 2015b; Guo et al., 2016; LeCun et al., 1990). Other approaches instead employ more sophisticated Bayesian estimators (Blundell et al., 2015; Graves, 2011; Nalisnick et al., 2015; Ullrich et al., 2017; Molchanov et al., 2017). However, these cases cannot significantly reduce computation times and memory without some special coding and processing because the dimensionality of the neurons/activations have not been changed.

To address the latter, it is necessary to target activations for pruning, which is our primary focus herein. For this purpose, multiple interesting Bayesian approaches have been proposed (Louizos et al., 2017a; Neklyudov et al., 2017) that rely on sparse priors (e.g., Jeffreys, horseshoe) on either groups of weights along dimensions useful for compression, or directly on the activations themselves. A variational free energy approximation/bound on the log-likelihood is then optimized, potentially using the warm-start procedure from (Sønderby et al., 2016) to gradually increase the influence of regularization effects (at the expense of altering the original bound). Although the final objectives minimized by these approaches significantly overlap with each other and our method, they are derived from a completely different perspective and involve several key, differentiating assumptions.

As an alternative deterministic strategy, (Liu et al., 2017; Pan et al., 2016; Wen et al., 2016) address similar pruning effects by applying convex group Lasso or $\ell_1$ norm-based regularizers in various different ways. In contrast, (Louizos et al., 2017b) first adopts an ideal $\ell_0$ norm penalty on rows or columns of weight matrices, and then deals with the resulting discontinuous, non-convex energy surface via a probabilistic smoothing approximation. Finally, (Lin et al., 2017) introduces a radically different approach based on reinforcement learning while (Li et al., 2016) specifically prunes convolutional neural network filters that have small norms.

## 3. Model Development

We first introduce some brief notational details. Denote the input variables to a neural network with $L$ layers as $\boldsymbol{x} \in \mathbb{R}^d$ and the associated label (or target output) as $\boldsymbol{y} \in \mathcal{Y}$. We represent the network hidden layer activations as $\{\boldsymbol{h}_i\}_{i=1}^L$, where $\boldsymbol{h}_i \in \mathbb{R}^{r_i}$.

Now if we view $\boldsymbol{x}$ as a stochastic input, feedforward network layers are sometimes interpreted as a Markov chain of successive representations (Tishby & Zaslavsky, 2015), *i.e.*,

$$\boldsymbol{y} \to \boldsymbol{x} \to \boldsymbol{h}_1 \to \ldots \to \boldsymbol{h}_L \to \hat{\boldsymbol{y}}. \tag{1}$$

Every hidden layer in the network defines the conditional probability $p(\boldsymbol{h}_i|\boldsymbol{h}_{i-1})$, where we use $\boldsymbol{x} = \boldsymbol{h}_0$ for convenience. For a deterministic network model, $p(\boldsymbol{h}_i|\boldsymbol{h}_{i-1})$ can be regarded as a Dirac-delta function; however, there are also many situations where the hidden layers are stochastic even when conditioned on their inputs. For example, when using dropout (Srivastava et al., 2014), each hidden neuron has some probability of being set to zero. Likewise, in Bayesian neural networks (Blundell et al., 2015) and variational autoencoder models (Rezende et al., 2014; Kingma & Welling, 2014) some or all layer activations are assigned a non-degenerate distribution. In most such cases, the neurons within each stochastic hidden layer are assumed to be conditionally independent when conditioned on the activations from the previous layer, *e.g.*, a diagonal Gaussian distribution.

The role of the hidden layers is to extract information from the previous ones, while the output layer attempts to approximate the true distribution $p(\boldsymbol{y}|\boldsymbol{h}_L)$ via some tractable alternative $q(\boldsymbol{y}|\boldsymbol{h}_L)$. However, even if sufficient information for accurately predicting $\boldsymbol{y}$ percolates through the network to the output, many of the internal representations may still contain superfluous content. Removal of this redundant content through some form of pruning or network ablation therefore represents a viable route to model compression.

Our starting point for accomplishing this goal is to explicitly penalize an information theoretic measure of redundancy between each adjacent layer, a concept originally introduced as the information bottleneck (Tishby et al., 2000). More specifically, for every hidden layer $\boldsymbol{h}_i$, we would like to *minimize* the mutual information $\boldsymbol{I}(\boldsymbol{h}_i;\boldsymbol{h}_{i-1})$ between $\boldsymbol{h}_i$ and $\boldsymbol{h}_{i-1}$ to remove inter-layer redundancy, while simultaneously *maximizing* the mutual information $\boldsymbol{I}(\boldsymbol{h}_i;\boldsymbol{y})$ between $\boldsymbol{h}_i$ and the output $\boldsymbol{y}$ to encourage accurate predictions of $\boldsymbol{y}$. Consequently, the layer-wise energy $\mathcal{L}_i$ becomes

$$\mathcal{L}_i = \gamma_i \boldsymbol{I}(\boldsymbol{h}_i;\boldsymbol{h}_{i-1}) - \boldsymbol{I}(\boldsymbol{h}_i;\boldsymbol{y}), \tag{2}$$

where $\gamma_i \geq 0$ is a coefficient that determines the strength of the bottleneck, or the degree to which we value compression over prediction accuracy. Summing over layers, the goal then is to minimize $\sum_i \mathcal{L}_i$ with respect to both network weights and any additional parameters describing the distributions $q(\boldsymbol{y}|\boldsymbol{h}_L)$ and $p(\boldsymbol{h}_i|\boldsymbol{h}_{i-1})$ for all $i$. However, reasonable model choices reflecting popular network architectures do not facilitate tractable computation of (2). Fortunately though, certain variational bounds can serve as a convenient surrogate. In this work, we invoke the upper bound

$$\tilde{\mathcal{L}}_i \;=\; \gamma_i \mathbb{E}_{\boldsymbol{h}_{i-1}\sim p(\boldsymbol{h}_{i-1})}[\mathbb{KL}[p(\boldsymbol{h}_i|\boldsymbol{h}_{i-1})||q(\boldsymbol{h}_i)]] - \mathbb{E}_{\{\boldsymbol{x},\boldsymbol{y}\}\sim\mathcal{D},\boldsymbol{h}\sim p(\boldsymbol{h}|\boldsymbol{x})}\left[\log q(\boldsymbol{y}|\boldsymbol{h}_L)\right] \;\geq\; \mathcal{L}_i, \tag{3}$$

where $\boldsymbol{h}_{1:i} \triangleq \{\boldsymbol{h}_j\}_{j=1}^{i}$, $\boldsymbol{h} \triangleq \boldsymbol{h}_{1:L}$, $\mathcal{D}$ denotes the true data distribution, and $q(\boldsymbol{h}_i)$ and $q(\boldsymbol{y}|\boldsymbol{h}_L)$ represent two variational distributions designed to approximate $p(\boldsymbol{h}_i)$ and $p(\boldsymbol{y}|\boldsymbol{h}_L)$ respectively. Details of the derivations can be found in Appendix A.

$\tilde{\mathcal{L}}_i$ from (3) is composed of two terms. The first is the KL divergence between $p(\boldsymbol{h}_i|\boldsymbol{h}_{i-1})$ and $q(\boldsymbol{h}_i)$, which approximates how much information $\boldsymbol{h}_i$ extracts from $\boldsymbol{h}_{i-1}$. The second term reflects fidelity with respect to the data distribution. The final variational information bottleneck loss function then becomes

$$\tilde{\mathcal{L}} \;\triangleq\; \sum_i \tilde{\mathcal{L}}_i \tag{4}$$

to assimilate information management across all layers.

Of course to actually optimize (4) we need to specify a parametric form for the distributions $p(\boldsymbol{h}_i|\boldsymbol{h}_{i-1})$, $q(\boldsymbol{h}_i)$, and $q(\boldsymbol{y}|\boldsymbol{h}_L)$. For the latter, the final network layer with weights $\boldsymbol{W}_y$ provides the necessary structure, often a multinomial distribution for classification problems and a Gaussian for standard regression tasks. And with respect to the conditional layer-wise distributions, we assume that each $p(\boldsymbol{h}_i|\boldsymbol{h}_{i-1})$ is defined via the relation

$$\boldsymbol{h}_i = (\boldsymbol{\mu}_i + \boldsymbol{\epsilon}_i \odot \boldsymbol{\sigma}_i) \odot f_i(\boldsymbol{h}_{i-1}), \tag{5}$$

where $\boldsymbol{\sigma}_i$ and $\boldsymbol{\mu}_i$ are learnable parameters and $\boldsymbol{\epsilon}_i$ is a random vector sampled from $\mathcal{N}(\boldsymbol{0},\boldsymbol{I})$. In contrast, the function $f_i$ represents a typical, deterministic network layer, meaning the concatenation of a linear transformation (or convolution operation), batch normalization, and some nonlinear activation function. In fact, if we were to fix $\boldsymbol{\mu}_i = \boldsymbol{1}$ and $\boldsymbol{\sigma}_i = \boldsymbol{0}$, then the model would default to a regular feed-forward neural network. Additionally, we use $\boldsymbol{W}_i$ to indicate the weights embedded in $f_i$. In addition, $\boldsymbol{W}_{i,j\cdot}$ represents the $j$-th row of $\boldsymbol{W}_i$ while $\boldsymbol{W}_{i,\cdot j}$ denotes the $j$-th column. Consequently, $\boldsymbol{W}_{i,j\cdot}$ corresponds with the $j$-th neuron in the $i$-th hidden layer, *i.e.*, $h_{i,j}$, and $\boldsymbol{W}_{i,\cdot j}$ corresponds to the $j$-th neuron in the $(i-1)$-th hidden layer, *i.e.* $h_{i-1,j}$. To avoid unnecessary clutter, we omit the bias in all layers.

With the above definitions in mind, it follows that

$$p(\boldsymbol{h}_i|\boldsymbol{h}_{i-1}) = \mathcal{N}\left(\boldsymbol{h}_i; f_i(\boldsymbol{h}_{i-1}) \odot \boldsymbol{\mu}_i, \mathrm{diag}[f_i(\boldsymbol{h}_{i-1})^2 \odot \boldsymbol{\sigma}_i^2]\right). \tag{6}$$

Note that Gaussian noise has previously been multiplied with layer-wise activations as a path towards network regularization (Kingma et al., 2015; Achille & Soatto, 2018); however, these works are not concerned with network compression and other modeling details are significantly different than ours. It has also been used in (Neklyudov et al., 2017) for neuron pruning in conjunction with a truncated, approximate Jeffreys prior. But this requires additional hyper-parameters for balancing the approximation, without which their alternative energy function is ill-defined. Moreover, the relationship between these parameters and compression performance is unclear.

Proceeding further, with our model we simply specify that $q(\boldsymbol{h}_i)$ is also Gaussian via

$$q(\boldsymbol{h}_i) = \mathcal{N}\left(\boldsymbol{h}_i; \boldsymbol{0}, \mathrm{diag}[\boldsymbol{\xi}_i]\right), \tag{7}$$

where $\boldsymbol{\xi}_i$ is an unknown vector of variances that can be learned from the data. Note that if any of these variances are pushed to zero during training, this action will in turn pressure the corresponding coordinates of $p(\boldsymbol{h}_i|\boldsymbol{h}_{i-1})$ towards a degenerate Dirac-delta, effectively pruning the associated neuron from the model. As we will later argue, both theoretically and empirically, this form of regularization can serve as a powerful basis for network compression and redundancy reduction.[1] And with regard to practical implementations, it is far easier to exploit network compression instantiated via neural pruning than by, for example, reducing the information content uniformly across all the neurons in a layer.

Our Gaussian assumptions are also advantageous in that they lead to an interpretable, closed-form approximation for the KL term from (3), allowing us to directly optimize $\boldsymbol{\xi}_i$ out of the model. Specifically, following several algebraic manipulations shown in Appendix B, we have that

$$\inf_{\boldsymbol{\xi}_i \succ \boldsymbol{0}} \; 2\mathbb{E}_{\boldsymbol{h}_{i-1}\sim p(\boldsymbol{h}_{i-1})}\left[\mathbb{KL}\left[p(\boldsymbol{h}_i|\boldsymbol{h}_{i-1})||q(\boldsymbol{h}_i)\right]\right] \;\equiv\; \sum_j \left[\log\left(1+\frac{\mu_{i,j}^2}{\sigma_{i,j}^2}\right) + \psi_{i,j}\right], \tag{8}$$

where

$$\psi_{i,j} \;\triangleq\; \log \mathbb{E}_{\boldsymbol{h}_{i-1}\sim p(\boldsymbol{h}_{i-1})}\left[f_{i,j}(\boldsymbol{h}_{i-1})^2\right] - \mathbb{E}_{\boldsymbol{h}_{i-1}\sim p(\boldsymbol{h}_{i-1})}\left[\log f_{i,j}(\boldsymbol{h}_{i-1})^2\right] \tag{9}$$

and $\mu_{i,j}$, $\sigma_{i,j}$, and $f_{i,j}(\boldsymbol{h}_{i-1})$ denote the $j$-th element of $\boldsymbol{\mu}_i$, $\boldsymbol{\sigma}_i$, and $f_i(\boldsymbol{h}_{i-1})$ respectively. The quantity $\psi_{i,j}$ is always positive by Jensen's inequality, but likely to be smaller when the variance of $p(\boldsymbol{h}_{i-1})$ is not too large and the gap between the log of an expectation and the expectation of the log is reduced. For computational convenience, and because empirically we found the contribution of $\psi_{i,j}$ to be unnecessary for excellent compression performance,

1. In the past a similar prior has been used in conjunction with learning sparse kernel machines (Tipping, 2001).

we remove this factor from the model. By plugging this simplified KL approximation into (3) across all layers $i$, we obtain the revised final loss function[2]

$$\tilde{\mathcal{L}} \;=\; \sum_{i=1}^{L} \gamma_i \sum_{j=1}^{r_i} \log\left(1 + \frac{\mu_{i,j}^2}{\sigma_{i,j}^2}\right) - L\; \mathbb{E}_{\{\boldsymbol{x},\boldsymbol{y}\}\sim\mathcal{D},\boldsymbol{h}\sim p(\boldsymbol{h}|\boldsymbol{x})}\left[\log q(\boldsymbol{y}|\boldsymbol{h}_L)\right], \tag{10}$$

where $r_i$ denotes the number of neurons or filters in the $i$-th layer.

Several items are worth pointing out with respect to this objective. First, the parameters $\gamma_i$ grant us the flexibility to individually tailor the degree of compression across each layer like some prior methods. While in many situations the simple choice $\gamma_i = \gamma > 0$ for all $i$ is sufficient, in cases where there are significant complexity differences across layers a simple modification can be warranted. Regardless, $\gamma_i$ serves a useful, transparent purpose, and our energy function requires no additional hyper-parameters as in (Louizos et al., 2017a,b; Neklyudov et al., 2017) to describe priors, approximations, or any other constraints.

Secondly, the weighting factor $L$ on the data term naturally provides balance for deeper networks, preventing the KL factors from accumulating such that the prediction accuracy is completely ignored by the globally optimal solution. In contrast, with many probabilistic network models, a related KL term must be heuristically down-weighted during training (Louizos et al., 2017a; Sønderby et al., 2016), but this then interferes with the associated free-energy bound on the log-likelihood. This is unlike our approach, where $L$ naturally emerges from the formulation itself, and represents an integral part of the variational information bottleneck bounding process.

And finally, although the remaining integrals from (10) have no closed form, unbiased stochastic approximations of the required expectations provide a natural remedy for training purposes (Kingma & Welling, 2014; Rezende et al., 2014). First, a pair $\{\boldsymbol{x}, \boldsymbol{y}\}$ is randomly sampled from the training data and fed into the network. For the forward pass, at each layer we sample $\boldsymbol{\epsilon}_i \sim \mathcal{N}(\boldsymbol{0}, \boldsymbol{I})$ and then compute $\boldsymbol{h}_i$ via (5). For the backward pass, the gradients can naturally flow via back-propagation to $\{\boldsymbol{\mu}_i, \boldsymbol{\sigma}_i, \boldsymbol{W}_i\}_{i=1}^L$ and $\boldsymbol{W}_y$.

We refer to our model as the *Variational Information Bottleneck Network* (VIBNet).[3] The layer-wise sampling strategy is shown in Figure 1. In subsequent sections we will provide supporting analyses that help to justify our choice of objective function.

## 4. Reduced Redundancy via Intrinsic Sparsity

In the previous section we motivated the VIBNet compression model using the concept of the information bottleneck. However, given that multiple bounds/approximations were

2. With slight abuse of notation, we reuse $\tilde{\mathcal{L}}$ to describe this updated objective even though in reality it is no longer a strict upper bound, satisfying only the looser requirement $\tilde{\mathcal{L}} \geq \sum_{i=1}^{L}\left(\mathcal{L}_i - \gamma_i \sum_{j=1}^{r_i} \psi_{i,j}\right)$. Note though that if the activation function is such that $f_{i,j}(\boldsymbol{h}_{i-1})^2 \approx 0$ across a region with nonzero probability measure, then the associated $\psi_{i,j}$ can potentially be arbitrarily large, trivializing the bound. Regardless, in later sections we will provide theoretical justification for adopting $\tilde{\mathcal{L}}$ that is independent of the tightness of this approximation anyway. Additionally, for simplicity and without loss of generality we have absorbed the factor of 2 from (8) into each $\gamma_i$.
3. The variational information bottleneck has been referenced in the past as a means of improving generalization performance and robustness to adversarial attacks (Alemi et al., 2016), but this is unrelated to our present purposes here.

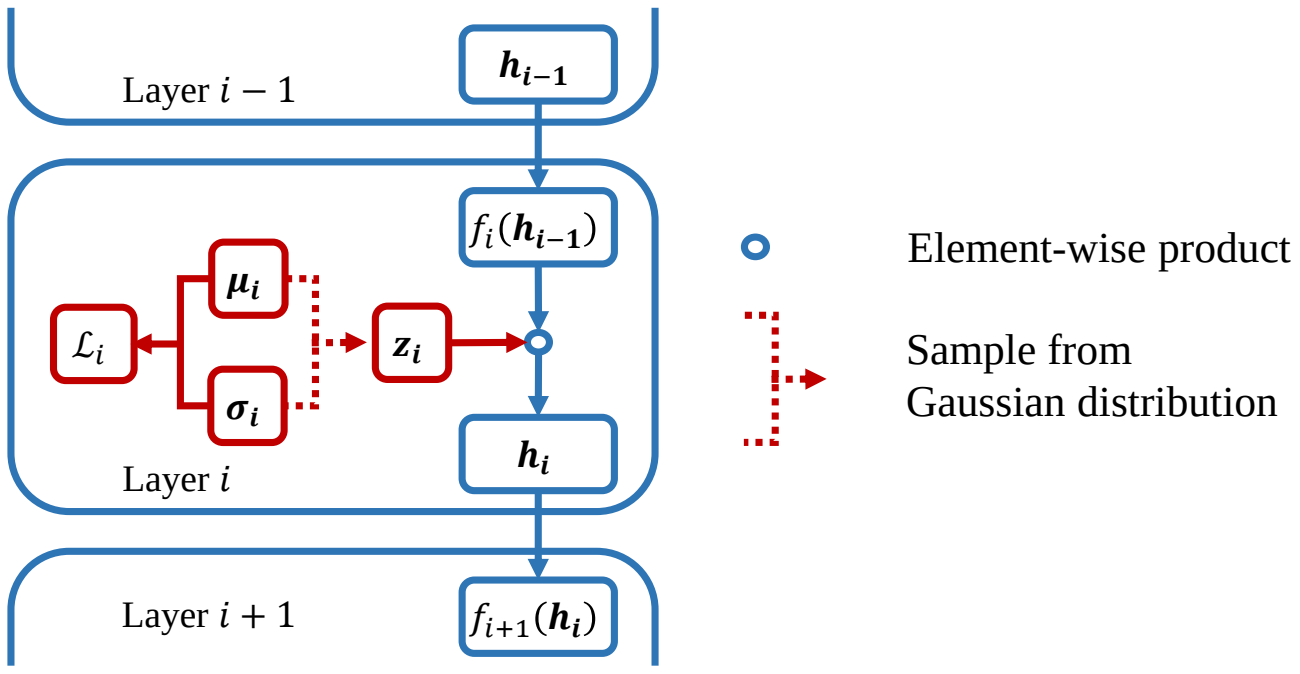

Figure 1: VIBNet Structure. The conditional distribution $p(\boldsymbol{h}_i|\boldsymbol{h}_{i-1})$ is given by (6). $\boldsymbol{h}_i$ is sampled by multiplying $f_i(\boldsymbol{h}_{i-1})$ with a random variable $\boldsymbol{z}_i \triangleq \boldsymbol{\mu}_i + \boldsymbol{\epsilon}_i \odot \boldsymbol{\sigma}_i$.

required to obtain a tractable energy function, it is quite reasonable to question whether or not our original design principles were somehow compromised. To address this concern, both this section and the sequel will attempt to provide independent justification for the VIBNet cost. In this way, we can naturally sidestep issues related to the tightness or legitimacy of the various underlying bounds involved in arriving at (10). From this high-level perspective, we can then view the information bottleneck as having merely provided a form of loose inspiration for a candidate energy function, but one that must still be further subject to careful examination before confidence is warranted.

To begin, recall that (10) is constructed from two factors: a regularizer based on the KL divergence, and a data fit term involving an expectation over latent hidden states. With respect to the former, it is easily shown that $\log(1+u)$ is a concave, non-decreasing function on the domain $[0,\infty)$, canonical characteristics of a sparsity-promotion regularizer (Chen et al., 2017). Therefore, rather than favoring a solution with many smaller, partially shrunken versions of the ratios

$$\alpha_{i,j} \triangleq \mu_{i,j}^2 \sigma_{i,j}^{-2}, \quad \forall i,j, \tag{11}$$

this type of sparsity archetype instead prefers pushing some percentage to exactly zero while leaving others mostly unchanged (Rao et al., 2003). But how exactly does this favoritism relate to our original information bottleneck criterion? We characterize this relationship with the following result:

**Proposition 1** *At any minimum of (10), $\alpha_{i,j} = 0$ is a necessary condition for $\boldsymbol{I}(h_{i,j};\boldsymbol{h}_{i-1}) = 0$ and a sufficient condition for $\boldsymbol{I}(h_{i,j};\boldsymbol{h}_{i-1}) \leq \psi_{i,j}$. Additionally, further assume that there exists a ball of radius $\rho > 0$ around a given minimum such that within this ball, the data term $-\mathbb{E}_{\{\boldsymbol{x},\boldsymbol{y}\}\sim\mathcal{D},\boldsymbol{h}\sim p(\boldsymbol{h}|\boldsymbol{x})}\left[\log q(\boldsymbol{y}|\boldsymbol{h}_L)\right]$ is an increasing function of $\sigma_{i,j} \geq 0$. Then $\alpha_{i,j} = 0$ is a sufficient condition for $\boldsymbol{I}(h_{i,j};\boldsymbol{h}_{i-1}) = 0$ at this minimum.*

Based on the data processing inequality (Cover & Thomas, 2012) and the Markovian structure from (1), $\boldsymbol{I}(h_{i,j};\boldsymbol{y}) \leq \boldsymbol{I}(h_{i,j};\boldsymbol{h}_{i-1})$. It directly follows that any neuron with $\boldsymbol{I}(h_{i,j};\boldsymbol{h}_{i-1}) = 0$ actually contains no information about $\boldsymbol{y}$ and is therefore redundant.

Hence such a neuron can be safely removed without hurting the predictive performance of the network. And if $\boldsymbol{I}(h_{i,j}; \boldsymbol{h}_{i-1}) = 0$, then $\alpha_{i,j} = 0$ per the necessary condition from Proposition 1. Consequently, if we remove all neurons with $\alpha_{i,j} = 0$, we can be sure that any survivors are informative, with $\boldsymbol{I}(h_{i,j}; \boldsymbol{h}_{i-1}) > 0$.

But these observations do not address the potential risk of over-pruning useful neurons. Fortunately though, if we manage to minimize the VIBNet cost and find that some $\alpha_{i,j} = 0$, then the sufficiency qualification of Proposition 1 implies that $\boldsymbol{I}(h_{i,j}; \boldsymbol{h}_{i-1}) = 0$, provided at least that, within an arbitrarily small neighborhood around this minimum, the data term $-\mathbb{E}_{\{\boldsymbol{x},\boldsymbol{y}\}\sim\mathcal{D},\boldsymbol{h}\sim p(\boldsymbol{h}|\boldsymbol{x})}[\log q(\boldsymbol{y}|\boldsymbol{h}_L)]$ is an increasing function of $\sigma_{i,j}$. This latter condition will provably hold in many special cases, e.g., if $\log q(\boldsymbol{y}|\boldsymbol{h}_L)$ is a quadratic function of the activations, but it is likely to be true in broader practical scenarios given that increasing the variance will generally be disruptive to the data fit, and hence increase the negative log-likelihood.

Of course if the weights from the next layer to which $h_{i,j}$ feeds are equal to zero, meaning $\boldsymbol{W}_{i+1,\cdot j} = \boldsymbol{0}$, then the stated data term can no longer be a strictly increasing function of $\sigma_{ij}$ within any ball no matter how small (just flat or non-decreasing). However, in this situation, even though technically $\boldsymbol{I}(h_{i,j}; \boldsymbol{h}_{i-1})$ may not be provably equal to zero at this minimum, it is irrelevant since no information $h_{i,j}$ retains about $\boldsymbol{h}_{i-1}$ can be passed on to the next layer. It is therefore still a useless neuron and should be pruned anyway.

This raises a larger point in terms of superfluous information unrelated to the relationship between $\boldsymbol{I}(h_{i,j}; \boldsymbol{h}_{i-1})$ and $\alpha_{i,j}$. In general, if a neuron contains some information pertaining to $\boldsymbol{y}$, but this information is never inherited by the next layer because $\boldsymbol{W}_{i+1,\cdot j} = \boldsymbol{0}$, then it is effectively useless and represents a prime candidate for compression. In this case, the corresponding $\alpha_{i,j}$ should *necessarily* also be zero if it is to serve as an ideal Bellwether for expendable neurons. Fortunately, this is indeed the case with our model:

**Proposition 2** *At any minimum of (10), $\alpha_{i,j} = 0$ is a necessary condition for $\boldsymbol{W}_{i+1,\cdot j} = \boldsymbol{0}$.*

Stated differently, this result implies that if $\boldsymbol{W}_{i+1,\cdot j} = \boldsymbol{0}$ and $\alpha_{i,j} \neq 0$, then we cannot be at a minimum of (10). And so again, the state $\alpha_{i,j} = 0$ is naturally aligned with our goal of unmasking ineffectual neurons for pruning.

At a high level then, given that our chosen penalty encourages $\alpha_{i,j} \to 0$, Propositions 1 and 2 then loosely suggest that this process may naturally aggregate useless information into certain expendable neurons, as opposed to distributing the bottleneck equally across all neurons in a layer. And this aggregation strategy provides a simple rule for exposing these redundant neurons such that they can be readily pruned from the model for practical compression purposes: namely, those neurons for which $\alpha_{i,j} \approx 0$ are ideal candidates for removal, and both of the potentially redundant neuron types described above will be discarded based on this criteria (providing further justification for drop-out (Molchanov et al., 2017)). We also conjecture that these relatively straightforward channels for reducing redundancy are indicative of broader mechanisms for compression.

However, there nonetheless remains a lingering issue with this overall line of reasoning. Although the KL-penalty should favor neuron pruning in a generic context as we have argued, the component factors $\boldsymbol{\mu}$ and $\boldsymbol{\sigma}$ are also nonlinearly combined within the neural-network-dependent data term. It therefore remains ambiguous exactly how the suggested sparsity mechanism will operate within this particular, practically-relevant setting. More-over, it is still unclear how the stochastic, sparsity-promoting objective of VIBNet may exhibit any advantage over standard, deterministic alternatives such as the use of $\ell_1$ norm or related penalties. We consider these issues next.

## 5. Analysis of Tractable Upper Bounds

In general, it is extremely difficult to analyze the complex energy surface of a deep network, and the problem is only compounded when we include the high-dimensional integrals from (10). Fortunately though, certain convenient bounds and analyses inspired by sparse Bayesian methods (Wipf et al., 2011) allow us to nonetheless gain insights into operational behaviors of VIBNet. At an intuitive level, the basic idea here is that if tractable upper bounds can reasonably describe a local neighborhood while displaying useful properties with respect to compression and neural pruning, then we may expect that the underlying energy itself may inherit these desirable attributes, at least to some extent in local regions well-matched to the bounds.

To begin, let $\boldsymbol{\theta} \triangleq \{\boldsymbol{\mu}_i, \boldsymbol{\sigma}_i\}_{i=1}^L$ and $\boldsymbol{W} \triangleq \left\{\{\boldsymbol{W}_i\}_{i=1}^L, \boldsymbol{W}_y\right\}$. We then define

$$g\left(\boldsymbol{\epsilon}; \boldsymbol{\theta}, \boldsymbol{W}\right) \triangleq -L \int p(\boldsymbol{x}, \boldsymbol{y}) \log q\left[\boldsymbol{y} | \boldsymbol{h}_L\left(\boldsymbol{\epsilon}, \boldsymbol{x}; \boldsymbol{\theta}, \boldsymbol{W}\right)\right] d\boldsymbol{x} d\boldsymbol{y}, \tag{12}$$

where the last hidden layer activation $\boldsymbol{h}_L\left(\boldsymbol{\epsilon}, \boldsymbol{x}; \boldsymbol{\theta}, \boldsymbol{W}\right)$ is described recursively via (5) and we have explicitly included its dependence on the random variables $\{\boldsymbol{\epsilon}, \boldsymbol{x}\}$, with $\boldsymbol{\epsilon} \triangleq \{\boldsymbol{\epsilon}_i\}_{i=1}^L$, and the parameters $\{\boldsymbol{\theta}, \boldsymbol{W}\}$. It then follows that the VIBNet objective from (10) can be re-expressed as

$$\tilde{\mathcal{L}}(\boldsymbol{\theta}, \boldsymbol{W}) = \int p(\boldsymbol{\epsilon}) g\left(\boldsymbol{\epsilon}; \boldsymbol{\theta}, \boldsymbol{W}\right) d\boldsymbol{\epsilon} + \sum_{i=1}^L \gamma_i \sum_{j=1}^{r_i} \log\left(1 + \alpha_{i,j}\right). \tag{13}$$

Note that we have included the parametric dependence of $\tilde{\mathcal{L}}$ on $\{\boldsymbol{\theta}, \boldsymbol{W}\}$ which serves to clarify certain usages later. Now suppose for any fixed $\boldsymbol{W} = \boldsymbol{W}'$ we construct a positive semi-definite quadratic upper bound on $g$ with respect to $\boldsymbol{z}(\boldsymbol{\epsilon}; \boldsymbol{\theta}) \triangleq \{\boldsymbol{z}_i(\boldsymbol{\epsilon}_i; \boldsymbol{\theta}_i)\}_{i=1}^L$ (stacked in vectorized form), with $\boldsymbol{z}_i(\boldsymbol{\epsilon}_i; \boldsymbol{\theta}_i) \triangleq \boldsymbol{\mu}_i + \boldsymbol{\sigma}_i \odot \boldsymbol{\epsilon}_i$. Specifically, this leads to the generic bound $\bar{g}\left(\boldsymbol{\epsilon}; \boldsymbol{\theta}\right) \geq g\left(\boldsymbol{\epsilon}; \boldsymbol{\theta}, \boldsymbol{W}'\right)$, where

$$\bar{g}\left(\boldsymbol{\epsilon}; \boldsymbol{\theta}\right) \triangleq \boldsymbol{z}(\boldsymbol{\epsilon}; \boldsymbol{\theta})^\top \boldsymbol{A}^\top \boldsymbol{A} \boldsymbol{z}(\boldsymbol{\epsilon}; \boldsymbol{\theta}) + \boldsymbol{b}^\top \boldsymbol{z}(\boldsymbol{\epsilon}; \boldsymbol{\theta}) + c, \tag{14}$$

and for simplicity we have ignored the implicit dependency of $\boldsymbol{A}$, $\boldsymbol{b}$, and $\boldsymbol{c}$ on the value of $\boldsymbol{W}'$. Such a bound is always possible, with $\bar{g}\left(\boldsymbol{\epsilon}; \boldsymbol{\theta}\right) = g\left(\boldsymbol{\epsilon}; \boldsymbol{\theta}, \boldsymbol{W}'\right)$ for at least some value(s) of $\boldsymbol{z}(\boldsymbol{\epsilon}; \boldsymbol{\theta})$ provided that, for example, $g\left(\boldsymbol{\epsilon}; \boldsymbol{\theta}, \boldsymbol{W}'\right)$ has Lipschitz continuous gradients. Moreover, (14) can likewise be used to bound the overall cost via

$$\begin{aligned} \bar{\mathcal{L}}(\boldsymbol{\theta}) &\triangleq \int p(\boldsymbol{\epsilon}) \bar{g}\left(\boldsymbol{\epsilon}; \boldsymbol{\theta}\right) d\boldsymbol{\epsilon} + \sum_{i=1}^L \gamma_i \sum_{j=1}^{r_i} \log\left(1 + \alpha_{i,j}\right) \\ &\geq \tilde{\mathcal{L}}(\boldsymbol{\theta}, \boldsymbol{W}') \end{aligned} \tag{15}$$

This leads to the following:

**Proposition 3** *If $\boldsymbol{\theta}^* = \{\boldsymbol{\mu}^*, \boldsymbol{\sigma}^*\}$ is a local minimum of the bound $\bar{\mathcal{L}}(\boldsymbol{\theta})$ from (15), then*

$$\|\boldsymbol{\mu}^*\|_0 = \|\boldsymbol{\alpha}^*\|_0 \leq rank[\boldsymbol{A}] + 1. \tag{16}$$

Here $\|\cdot\|_0$ denotes the $\ell_0$ (quasi)-norm, or a count of the number of nonzero elements in a vector. It is typically viewed as the canonical or ideal measure of sparse solutions (Donoho & Elad, 2003). In general, Proposition 3 only provides a non-trivial upper bound on the estimated sparsity of $\boldsymbol{\mu}$ and the ratios $\boldsymbol{\alpha} = \boldsymbol{\mu}^2 \odot \boldsymbol{\sigma}^{-2}$ when $\boldsymbol{A}^\top \boldsymbol{A}$ is rank deficient, or equivalently, $\text{rank}[\boldsymbol{A}] < \dim[\boldsymbol{\mu}]$. However, given an overparameterized neural network where significant compression is possible, we expect that many regions of the energy landscape will be heavily skewed with long valleys of constant cost such that a low rank $\boldsymbol{A}$ contributes to a reasonable local approximation. So in this situation the upper bound indicates that some neuron pruning will occur even if only the worst local minimizer is obtained. Of course in practice far more significant sparsity is likely.

But there is another more subtle benefit of the VIBNet pruning mechanism: loosely speaking, *the shape/concavity of the implicit VIBNet sparsification effect is automatically calibrated with the local curvature of $g(\boldsymbol{\epsilon}; \boldsymbol{\theta}, \boldsymbol{W})$ in such a way that the ideal $\ell_0$ norm can be adaptively approximated while reducing the risk of bad local minimum.* To better appreciate this claim, it is helpful to first consider a more traditional, deterministic sparsity-based regularization analogue.

Suppose we were to remove the stochastic elements from the bound defined in (15), meaning we set $\boldsymbol{\sigma} = \boldsymbol{0}$, and we then replaced the KL-based penalty with some generic function $\pi$ promoting sparse values of $\boldsymbol{\mu}$. The result would be the deterministic energy

$$\Psi_\pi(\boldsymbol{\mu}) \triangleq \boldsymbol{\mu}^\top \boldsymbol{A}^\top \boldsymbol{A} \boldsymbol{\mu} + \boldsymbol{b}^\top \boldsymbol{\mu} + \sum_{i=1}^{L} \gamma_i \sum_{j=1}^{r_i} \pi(\mu_{i,j}). \tag{17}$$

If $\pi(\mu_{i,j})$ is the non-convex indicator function $\mathcal{I}[\mu_{i,j} \neq 0]$, then a weighted $\ell_0$ norm emerges; however, minimizing $\Psi_\pi(\boldsymbol{\mu})$ to obtain a maximally sparse solution is NP-hard because of a combinatorial number of local minima (likewise for smooth yet non-convex approximations (Chen et al., 2017)). In contrast, if $\pi(\mu_{i,j}) = |\mu_{i,j}|$, then we obtain a weighted $\ell_1$ norm regularizer, which represents the tightest convex relaxation of the $\ell_0$ norm. While the overall energy is now convex, minimization will often fail to produce maximally sparse (or maximally compressible) estimates except in highly idealized scenarios (Donoho & Elad, 2003). This is because the $\ell_1$ norm tends to over-shrink large coefficients at the expense of sparsity (Fan & Li, 2001).

Against this backdrop, we can directly contrast the adaptive, data-dependent VIBNet regularization effect. Let $a_{i,j}$ denote the diagonal element of $\boldsymbol{A}^\top \boldsymbol{A}$ corresponding with $\mu_{i,j}$ and define $\omega_{i,j} \triangleq \gamma_i a_{i,j}^{-1}$. Then based on the proof of Proposition 3, it can be shown that bound $\bar{\mathcal{L}}(\boldsymbol{\theta})$ satisfies

$$\inf_{\boldsymbol{\sigma} \succ \boldsymbol{0}} \bar{\mathcal{L}}(\boldsymbol{\theta}) = \boldsymbol{\mu}^\top \boldsymbol{A}^\top \boldsymbol{A} \boldsymbol{\mu} + \boldsymbol{b}^\top \boldsymbol{\mu} + \sum_{i=1}^{L} \gamma_i \sum_{j=1}^{r_i} \rho(\mu_{i,j}; \omega_{i,j}), \tag{18}$$

where

$$\rho(\mu; \omega) \triangleq \frac{2|\mu|}{|\mu| + \sqrt{\mu^2 + 4\omega}} + \log\left(2\omega + \mu^2 + |\mu|\sqrt{\mu^2 + 4\omega}\right). \tag{19}$$

Functions like $\rho$ have previously been used for blind image deblurring (Wipf & Zhang, 2014), in part because they provide an attractive means of interpolating between scaled versions of the $\ell_0$ norm as $\omega \to 0$, and the $\ell_1$ norm as $\omega \to \infty$. This interpolation can, for example, be useful in adapting to different blur kernel estimates. Of paramount importance though, in the present context here this interpolation is directly modulated by the parameters $a_{i,j}$, which collectively represent a measure of the local curvature of $\bar{g}(\boldsymbol{\epsilon}; \boldsymbol{\theta})$ when $\boldsymbol{\sigma} = \mathbf{0}$, *i.e.*, a proxy for the deterministic, data-dependent neural network loss after the stochastic hidden layer latent variables have been removed.

The cumulative effect is that the penalty function shape is roughly matched to this data term. When the latter is relatively smooth and unconstrained, many $a_{i,j}$ values become small within the most representative bound. This pushes the corresponding $\omega_{i,j}$ to be large, and the regularizer is likewise comparably smooth and flat. This helps to avoid aggressive or premature dominance of a highly non-convex sparsity penalty in regions where the deep network energy is relatively flat.

Conversely, if the network's local region is highly curved and constrained, the bound reflecting local curvature will have many $a_{i,j}$ values that are large, the associated $\omega_{i,j}$ then becomes small, and a more $\ell_0$-norm-like regularizer emerges. But here a stronger penalty can be employed with relatively limited risk of a quick, dominant descent to far away spurious optima. For these reasons, we believe that such a data-dependent, adaptive regularizer is particularly appropriate for compression purposes.

## 6. Experiments and Discussion

In the majority of recent neural network compression work, models are evaluated with respect to some subset of the following architecture/dataset combinations: LeNet-300-100 (LeCun et al., 1998) and LeNet-5-Caffe[4] networks on MNIST (LeCun, 1998), and VGG-16 (Simonyan & Zisserman, 2014)[5] networks on CIFAR10 and CIFAR100 (Krizhevsky & Hinton, 2009). Using all of these benchmarks, we compare our VIBNet with published results from a variety of contemporary state-of-the-art methods including *Generalized Dropout* (GD) (Srinivas & Babu, 2016), *Group Lasso* (GL) (Wen et al., 2016), *Sparse Variational Dropout* (VD) (Molchanov et al., 2017), *Bayesian Compression with Group Normal Jeffreys Prior* (BC-GNJ) and *Group Horseshoe Prior* (BC-GHS) (Louizos et al., 2017a), *Sparse $\ell_0$ Regularization* (L0) and L0 with *separate* $\lambda$ for each layer (L0-sep) (Louizos et al., 2017b), *Drop Neuron* (DN) (Pan et al., 2016), *Runtime Neural Pruning* (RNP) (Lin et al., 2017), *Pruning Filter* (PF) (Li et al., 2016), *Network Slimming* (NS) (Liu et al., 2017), and *Structured Bayesian Pruning* (SBP) and SBP with *KL scaling* (SBPa) (Neklyudov et al., 2017).

Note that GL, DN, and NS all rely on an $\ell_1$-norm-like group-sparsity penalty in some form, and therefore may be at least partially exposed to some of the weaknesses described in Section 5. Likewise, L0-sep is based on a smoothed version of the $\ell_0$-norm obtained via an expectation operator over an additional set of latent variables, and therefore represents another interesting comparison. We also emphasize that because many of the above methods

4. https://github.com/BVLC/caffe/tree/master/examples/mnist
5. The original VGG-16 is applied on $224 \times 224$ images. Modified VGG16 versions are used in our experiments.

were introduced concurrently, none of them are actually compared against all of the others on standard benchmarks as we do here.

**Evaluation Metrics:** Beyond classification error on test sets, we also evaluate with respect to three metrics that relate to the compression ratio and model complexity:

1. *Model size* ($r_W$) - The ratio of number of nonzero weights in the compressed network versus the original model.

2. *Floating point operations* (FLOPs) - The number of floating point operations required to predict $\boldsymbol{y}$ from $\boldsymbol{x}$ during test-time.[6]

3. *Run-time memory footprint* ($r_N$) - The ratio of the space for storing hidden feature maps during run-time in the pruned network versus the original model. This involves calculating the feature map sizes (product of the channel, height, and width) across each layer.

**Training:** Our energy function only has the single parameter vector $\boldsymbol{\gamma}$ that balances compression versus accuracy across each layer. For LeNet-300-100 we simply use $\gamma_i = \gamma'$, *i.e.*, a constant for all layers. For LeNet-5-Caffe, we followed the approach of (Louizos et al., 2017b), which has a related layer-wise parameter. In contrast, for the larger VGG networks, we choose $\gamma_i = \gamma'/S_i$, where $S_i$ is the side length of the feature maps in the convolutional layers, and one for the fully connected layers. This simple rule, similar to a strategy from (Louizos et al., 2017b; Neklyudov et al., 2017), helps to account for different layer sizes, and the scalar $\gamma'$ remains the only tuning parameter. Overall, to best calibrate with prior methods, we chose $\gamma'$ to roughly match the accuracy of the best previously reported result. In this way if the resulting compression is superior, then we have a convincing unambiguous advantage. Otherwise, for arbitrary choices of $\gamma'$, clear comparisons are difficult if, for example, the accuracy is worse but the compression is much better. We also applied batch normalization (Ioffe & Szegedy, 2015) and weight decay to accelerate and regularize the training process, consistent with prior approaches.

As with other methods, we prune the VIBNet neurons after training whenever $\alpha_{i,j}$ is sufficiently small, consistent with Proposition 1. This is because VIBNet is trained stochastically and it is impossible for $\alpha_{i,j}$ to become exactly zero. We chose a simple hard-threshold for all experiments; however, we found that performance with respect to both accuracy and compression was insensitive to this choice since there is generally a clear separation between redundant and informative neurons. At this point, we may also further fine-tune the resulting compressed network weights as in (Li et al., 2016; Liu et al., 2017) to boost the final accuracy if desired (this is relatively efficient anyway since the network is now much smaller). Unless explicitly noted, however, no fine-tuning was used.

**Testing:** In the test phase, we only use the mean values of $p(\boldsymbol{h}_i|\boldsymbol{h}_{i-1})$ rather than sampling, which is computationally expensive. Hence we are ultimately only using a probabilistic network structure and the information bottleneck as a means of obtaining what can be viewed as a useful energy function for compressing what amounts to a deterministic

6. We count each multiplication as a single FLOP and exclude additions since typically #-multiplies = #-additions, consistent with most prior work. But for consistent comparisons here, we convert all alternative FLOP count schemes to this standard format.

| Method | $r_W(\%)$ | $r_N(\%)$ | error(%) | Pruned Model |
|---|---|---|---|---|
| VD | 25.28 | 58.95 | 1.8 | 512-114-72 |
| BC-GNJ | 10.76 | 32.85 | 1.8 | 278-98-13 |
| BC-GHS | 10.55 | 34.71 | 1.8 | 311-86-14 |
| L0 | 26.02 | 45.02 | **1.4** | 219-214-100 |
| L0-sep | 10.01 | 32.69 | 1.8 | 266-88-33 |
| DN | 23.05 | 57.94 | 1.8 | 542-83-61 |
| VIBNet | **3.59** | **16.98** | 1.6 | **97-71-33** |

Table 1: Compression results on MNIST using LeNet-300-100. VIBNet achieves much better compression than all previous methods while the error rate is nearly the best.

network. But certainly the option remains for sampling to improve accuracy as suggested in (Louizos et al., 2017a).

### 6.1 MNIST Results with LeNet-300-100 and LeNet-5

Perhaps the most commonly used pipeline for evaluating existing compression algorithms is MNIST hand-written image data paired with either the LeNet-300-100 or LeNet-5-Caffe network architecture. In evaluations, we follow the conventional training and testing protocols, initializing the weights from scratch like most methods applied to this data. Also since LeNet-300-100 is a fully connected network treating the input as an abstract vector rather than a 2D image, it makes sense to add an additional information bottleneck to the input layer.

Results for available models are shown in Table 1, where VIBNet achieves a much smaller $r_W$ and $r_N$ while the error rates for all methods are nearly the same. And the marginal 0.2 accuracy advantage of L0 is offset by the worse compression.

Next we evaluate VIBNet on the LeNet-5-Caffe network, which includes two convolutional layers and two fully connected layers. Results are shown in Table 2. Although VIBNet does not have the lowest $r_W$ (it is the second best in the table), it achieves the lowest FLOP and $r_N$. The accuracy is almost the same for all methods.

### 6.2 CIFAR10 and CIFAR100 Results Using VGG-16

Evaluations with larger VGG-16 networks on real-world CIFAR10 and CIFAR100 data are complicated by several factors. The primary issue is that, unlike MNIST, it becomes necessary to disentangle various sources of variation unrelated to compression algorithms. For example, different compression pipelines alter network structures and the form of the training data (e.g., data augmentation). Given then that there is no accepted standard for comparison, we evaluate against each competing pipeline individually, training and testing following each different published protocol. Additionally, in all cases models are initialized from a pre-trained network. Hence we obtain three separate VIBNet results for CIFAR10, and two separate results for CIFAR100.

We stress however, that VIBNet can work well under diverse conditions, including training from scratch instead of from a pre-trained network, and our optimal performance (which can be application-dependent) may not ultimately be represented by any of these training protocols originally developed for other models. Nonetheless, VIBNet still obtains uniformly improved results in all cases, which speaks to its versatility.

| Method | $r_W$(%) | FLOP(Mil) | $r_N$(%) | error(%) |
|---|---|---|---|---|
| GD | 1.38 | 0.250 | 32.00 | 1.1 |
| GL | 23.69 | 0.201 | 19.35 | 1.0 |
| VD | 9.29 | 0.660 | 60.78 | 1.0 |
| SBP | 19.66 | 0.213 | 21.15 | **0.9** |
| BC-GNJ | 0.95 | 0.283 | 35.03 | 1.0 |
| BC-GHS | **0.64** | 0.153 | 22.80 | 1.0 |
| L0 | 8.92 | 1.113 | 85.82 | **0.9** |
| L0-sep | 1.08 | 0.389 | 40.36 | 1.0 |
| VIBNet | 0.83 | **0.094** | **15.55** | 1.0 |

Table 2: Compression results on MNIST using Lenet-5-Caffe. VIBNet has the smallest FLOPs and $r_N$, while its $r_W$ is the second best. All methods achieve similar accuracy.

Table 3 displays separate VIBNet results against BC-GNJ/BC-GHS, PF/SBP/SBPa, and NS-Single/NS-Best following each separate protocol in turn. For the BC models, the dimension of the fully connected layers is simply changed from 4096 to 512, while leaving convolutional layers unaltered (Louizos et al., 2017a). PF and SBP/SBPa further remove one additional fully-connected layer (Li et al., 2016) and apply standard CIFAR10 data augmentation (e.g., cropping and flipping). Finally, the NS model replaces two fully connected layers with three convolutional layers, which can improve accuracy at the expense of FLOPs and model size (Liu et al., 2017); data augmentation is also used.

From the Table 3, VIBNet achieves the best performance across all three compression metrics, and comparable or better accuracy as well. SBPa achieves the second best compression, but this comes at the significant cost of a nearly 50% decrease in accuracy. Note that the NS algorithm involves multiple iterations of training and pruning; however, if this procedure is carried out too far, the accuracy drops significantly as the compression increases. Hence NS-Best refers to the iteration result with the best accuracy, while NS-Single refers to the first iteration, a comparable training process to VIBNet. Additionally, multiple iterations of training can be tedious in practice, especially since we do not know in advance how many iterations will be necessary. In contrast, lower accuracy with higher compression can be achieved via VIBNet by simply increasing $\gamma$.

Lastly, we compare performance on CIFAR100 against RNP and NS, where RNP (Lin et al., 2017) applies a similar network adaptation as PF on CIFAR10 without data augmentation, and NS is as described above. Result are shown in Table 4, with VIBNet showing consistent improvements.

| Method | $r_W$(%) | FLOP(Mil) | $r_N$(%) | error(%) |
|---|---|---|---|---|
| BC-GNJ | 6.57 | 141.5 | 81.68 | 8.6 |
| BC-GHS | 5.40 | 121.9 | 74.82 | 9.0 |
| VIBNet | **5.30** | **70.63** | **49.57** | 8.8 (**8.5**) |
| PF | 35.99 | 206.3 | 83.97 | 6.6 |
| SBP | 7.01 | 136.0 | 80.72 | 7.5 |
| SBPa | 5.78 | 99.20 | 66.46 | 9.0 |
| VIBNet | **5.45** | **86.82** | **57.86** | 6.5 (**6.1**) |
| NS-Single | 11.50 | 195.5 | - | 6.2 |
| NS-Best | 8.60 | 147.0 | - | 5.9 |
| VIBNet | **5.79** | **116.0** | **59.60** | 6.2 (**5.8**) |

Table 3: Compression results on CIFAR10 using VGG-16. We compare VIBNet to three different methods, in each case adopting the the training protocols of the original work. Although accuracy measures are similar, VIBNet produces the best compression by a significant margin. Error rates in parentheses were obtained by fine-tuning the pruned architecture. Note also that NS-Best involves multiple iterations of training and fine-tuning.

| Method | $r_W$(%) | FLOP(Mil) | $r_N$(%) | error(%) |
|---|---|---|---|---|
| RNP | - | 160 | - | 38.0 |
| VIBNet | **22.75** | **133.6** | **59.80** | 37.6 (**37.4**) |
| NS-Single | 24.90 | 250.5 | - | 26.5 |
| NS-Best | 20.80 | 214.8 | - | 26.0 |
| VIBNet | **15.08** | **203.1** | **73.80** | 25.9 (**25.7**) |

Table 4: Compression results on CIFAR100 using VGG-16. VIBNet is compared with two different protocols adopted from the corresponding prior work. Again, while accuracy measures are similar, VIBNet produces the best compression by a significant margin. The error rate in parentheses was obtained by fine-tuning the pruned architecture.

### 6.3 Redundancy Reduction Example

Finally, for illustration purposes we shift gears and examine how the intrinsic sparsity mechanism of VIBNet contributes to a layer-wise decrease in mutual information, consistent with the variational information bottleneck. To this end, we compute empirical estimates during training and compare against a traditional network. Of course in a deterministic model, the mutual information is infinite if we do not add noise. Therefore, to avoid making heuristic noise assumptions, we instead apply the non-parameteric mutual information estimator from (Kraskov et al., 2004) widely used for similar purposes, and compare against a single sample from VIBNet for fair comparison. Figure 2 shows that the mutual information estimates between the first hidden layer and input layer of LeNet-300-100. This value increases in the first several epochs and then starts to decrease once the VIBNet begins to compress the network. Other layers and networks behave similarly.

### 6.4 Discussion

We have compared our approach against arguably the largest number of competing methods primarily designed to prune neurons for computational and memory efficiency, achieving comparable accuracy with improved compression.

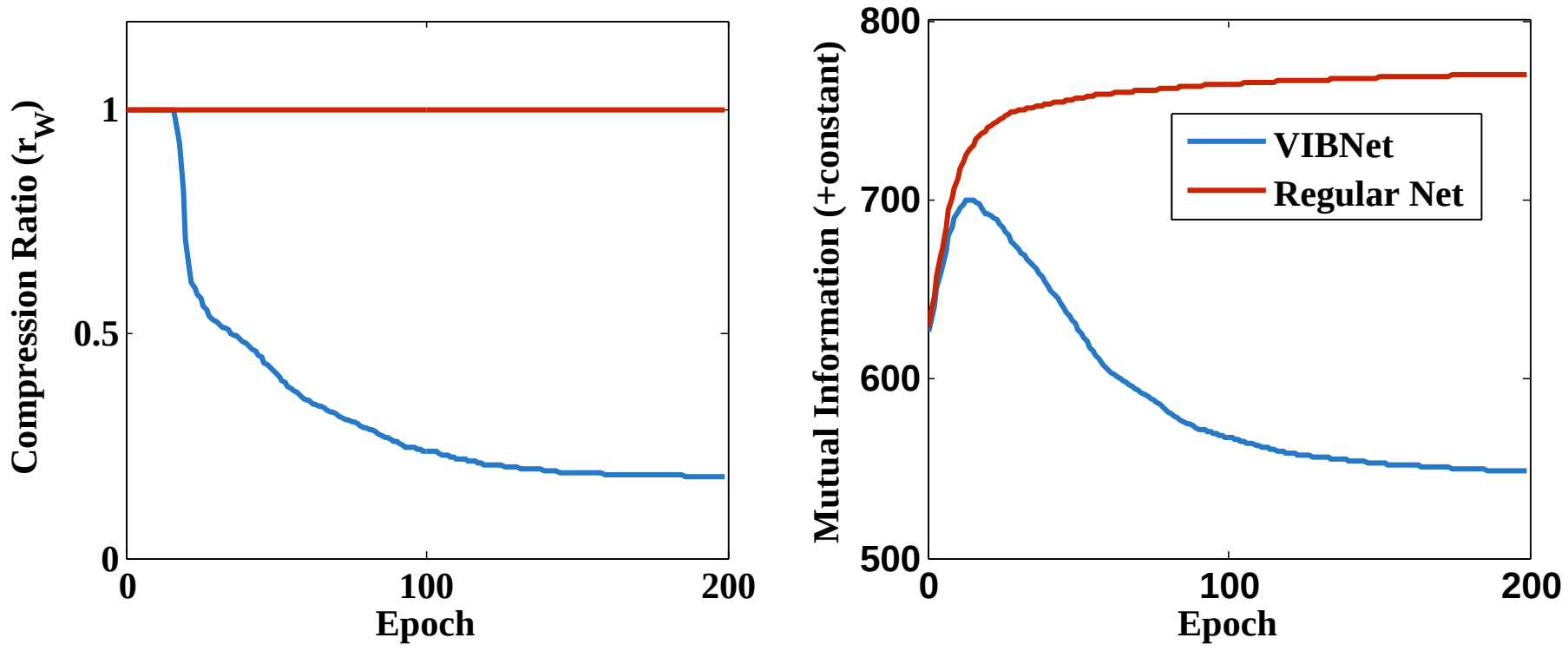

Figure 2: Mutual Information between $\boldsymbol{h}_1$ and $\boldsymbol{x}$ on LeNet-300-100. With a regular network, the mutual information always increases while the compression ratio stays at 1. In contrast, the mutual information of VIBNet increases in the first several epochs as the network attempts to learn a basic predictive model. Later when it starts to compress the network, the mutual information begins to decrease significantly.

## Appendix A. Derivation of the Variational Upper Bound (3)

The information bottleneck objective in (2) can be expressed as

$$\begin{aligned}
\mathcal{L}_i &= \gamma_i \boldsymbol{I}(\boldsymbol{h}_i;\boldsymbol{h}_{i-1}) - \boldsymbol{I}(\boldsymbol{h}_i;\boldsymbol{y}) \\
&= \int p(\boldsymbol{h}_i,\boldsymbol{h}_{i-1},\boldsymbol{y}) \left[\gamma_i \log \frac{p(\boldsymbol{h}_i,\boldsymbol{h}_{i-1})}{p(\boldsymbol{h}_i)p(\boldsymbol{h}_{i-1})} - \log \frac{p(\boldsymbol{h}_i,\boldsymbol{y})}{p(\boldsymbol{h}_i)p(\boldsymbol{y})}\right] d\boldsymbol{h}_i\boldsymbol{h}_{i-1}d\boldsymbol{y} \\
&= \int p(\boldsymbol{h}_i,\boldsymbol{h}_{i-1},\boldsymbol{y}) \left[\gamma_i \log p(\boldsymbol{h}_i|\boldsymbol{h}_{i-1}) - \gamma_i \log p(\boldsymbol{h}_i) - \log p(\boldsymbol{y}|\boldsymbol{h}_i) + \log p(\boldsymbol{y})\right] d\boldsymbol{h}_i\boldsymbol{h}_{i-1}d\boldsymbol{y} \\
&\equiv \int p(\boldsymbol{h}_i,\boldsymbol{h}_{i-1},\boldsymbol{y}) \left[\gamma_i \log p(\boldsymbol{h}_i|\boldsymbol{h}_{i-1}) - \gamma_i \log p(\boldsymbol{h}_i) - \log p(\boldsymbol{y}|\boldsymbol{h}_i)\right] d\boldsymbol{h}_i\boldsymbol{h}_{i-1}d\boldsymbol{y} \\
&\leq \int p(\boldsymbol{h}_i,\boldsymbol{h}_{i-1},\boldsymbol{y}) \left[\gamma_i \log p(\boldsymbol{h}_i|\boldsymbol{h}_{i-1}) - \gamma_i \log q(\boldsymbol{h}_i) - \log q(\boldsymbol{y}|\boldsymbol{h}_i)\right] d\boldsymbol{h}_i\boldsymbol{h}_{i-1}d\boldsymbol{y} \\
&= \int p(\boldsymbol{h}_i,\boldsymbol{h}_{i-1},\boldsymbol{y}) \left[\gamma_i \log \frac{p(\boldsymbol{h}_i|\boldsymbol{h}_{i-1})}{q(\boldsymbol{h}_i)} - \log q(\boldsymbol{y}|\boldsymbol{h}_i)\right] d\boldsymbol{h}_i\boldsymbol{h}_{i-1}d\boldsymbol{y} \qquad (20)\\
&= \int p(\boldsymbol{h}_{1:i},\boldsymbol{x},\boldsymbol{y}) \left[\gamma_i \log \frac{p(\boldsymbol{h}_i|\boldsymbol{h}_{i-1})}{q(\boldsymbol{h}_i)} - \log q(\boldsymbol{y}|\boldsymbol{h}_i)\right] d\boldsymbol{h}_{1:i}d\boldsymbol{x}d\boldsymbol{y} \\
&= \mathbb{E}_{\{\boldsymbol{x},\boldsymbol{y}\}\sim\mathcal{D},\boldsymbol{h}_{1:i-1}\sim p(\boldsymbol{h}_{1:i-1}|\boldsymbol{x})} \left[\int p(\boldsymbol{h}_i|\boldsymbol{h}_{i-1}) \left(\gamma_i \log \frac{p(\boldsymbol{h}_i|\boldsymbol{h}_{i-1})}{q(\boldsymbol{h}_i)} - \log q(\boldsymbol{y}|\boldsymbol{h}_i)\right) d\boldsymbol{h}_i\right],
\end{aligned}$$

where the equivalence in the fourth row comes from omitting the constant $\int p(\boldsymbol{y})\log p(\boldsymbol{y})d\boldsymbol{y}$ and the inequality in the fifth row comes from the Jesen's inequality. Now consider the factors inside the expectation. The first is the KL divergence between $p(\boldsymbol{h}_i|\boldsymbol{h}_{i-1})$ and $q(\boldsymbol{h}_i)$, which can be expressed either analytically or stochastically. Now we derive the second term. We assume

$$\begin{aligned}
q(\boldsymbol{y}|\boldsymbol{h}_i) &= \int q(\boldsymbol{y},\boldsymbol{h}_{i+1:L}|\boldsymbol{h}_i)d\boldsymbol{h}_{i+1:L} \qquad (21)\\
&= \int p(\boldsymbol{h}_{i+1:L}|\boldsymbol{h}_i)q(\boldsymbol{y}|\boldsymbol{h}_L)d\boldsymbol{h}_{i+1:L} \\
&= \int p(\boldsymbol{h}_{i+1}|\boldsymbol{h}_i)...p(\boldsymbol{h}_L|\boldsymbol{h}_{L-1})q(\boldsymbol{y}|\boldsymbol{h}_L)d\boldsymbol{h}_{i+1:L}.
\end{aligned}$$

Then we have

$$\begin{aligned}
&\int p(\boldsymbol{h}_i|\boldsymbol{h}_{i-1}) \log q(\boldsymbol{y}|\boldsymbol{h}_i)d\boldsymbol{h}_i \\
&= \mathbb{E}_{\boldsymbol{h}_i\sim p(\boldsymbol{h}_i|\boldsymbol{h}_{i-1})}\left[\log \int p(\boldsymbol{h}_{i+1:L}|\boldsymbol{h}_i)q(\boldsymbol{y}|\boldsymbol{h}_L)d\boldsymbol{h}_{i+1:L}\right] \\
&\geq \mathbb{E}_{\boldsymbol{h}_i\sim p(\boldsymbol{h}_i|\boldsymbol{h}_{i-1})}\left[\int p(\boldsymbol{h}_{i+1:L}|\boldsymbol{h}_i)\log q(\boldsymbol{y}|\boldsymbol{h}_L)d\boldsymbol{h}_{i+1:L}\right] \\
&= \mathbb{E}_{\boldsymbol{h}_i\sim p(\boldsymbol{h}_i|\boldsymbol{h}_{i-1})}\mathbb{E}_{\boldsymbol{h}_{i+1:L}\sim p(\boldsymbol{h}_{i+1:L}|\boldsymbol{h}_i)}\left[\log q(\boldsymbol{y}|\boldsymbol{h}_L)\right]. \qquad (22)
\end{aligned}$$

So the final upper bound of $\mathcal{L}_i$ becomes

$$\begin{aligned}
\mathcal{L}_i \leq\ & \gamma_i \mathbb{E}_{\{\boldsymbol{x},\boldsymbol{y}\}\sim\mathcal{D},\boldsymbol{h}_{1:i-1}\sim p(\boldsymbol{h}_{1:i-1}|\boldsymbol{x})}\left[\mathbb{KL}\left[p(\boldsymbol{h}_i|\boldsymbol{h}_{i-1})||q(\boldsymbol{h}_i)\right]\right] \\
& - \mathbb{E}_{\{\boldsymbol{x},\boldsymbol{y}\}\sim\mathcal{D},\boldsymbol{h}_{1:L}\sim p(\boldsymbol{h}_{1:L}|\boldsymbol{x})}\left[\log q(\boldsymbol{y}|\boldsymbol{h}_L)\right], \qquad (23)
\end{aligned}$$

which is the same as (3).

## Appendix B. KL Term Derivation

After plugging (6) and (7) into the KL term (8) and applying the standard formula for the KL divergence between two Gaussians, we obtain

$$\begin{aligned} & 2\mathbb{E}_{\boldsymbol{h}_{i-1}\sim p(\boldsymbol{h}_{i-1})}\left[\mathbb{KL}\left[p(\boldsymbol{h}_i|\boldsymbol{h}_{i-1})||q(\boldsymbol{h}_i)\right]\right] \\ = & \ \mathbb{E}_{\boldsymbol{h}_{i-1}\sim p(\boldsymbol{h}_{i-1})}\left[\sum_j \frac{\left(\mu_{i,j}^2+\sigma_{i,j}^2\right)\cdot f_{i,j}(\boldsymbol{h}_{i-1})^2}{\xi_{i,j}} - \log\frac{\sigma_{i,j}^2\cdot f_{i,j}(\boldsymbol{h}_{i-1})^2}{\xi_{i,j}} - 1\right]. \end{aligned} \tag{24}$$

To find the optimal $\xi_{i,j}$, denoted $\xi_{i,j}^*$, we take the gradient and set it equal to zero, giving us that

$$\mathbb{E}_{\boldsymbol{h}_{i-1}\sim p(\boldsymbol{h}_{i-1})}\left[-\frac{\left(\mu_{i,j}^2+\sigma_{i,j}^2\right)\cdot f_{i,j}(\boldsymbol{h}_{i-1})^2}{{\xi_{i,j}^*}^2} + \frac{1}{\xi_{i,j}^*}\right] = 0. \tag{25}$$

Solving this equation we obtain

$$\xi_{i,j}^* = \mathbb{E}_{\boldsymbol{h}_{i-1}\sim p(\boldsymbol{h}_{i-1})}\left[\left(\mu_{i,j}^2+\sigma_{i,j}^2\right)\cdot f_{i,j}(\boldsymbol{h}_{i-1})^2\right] = \left(\mu_{i,j}^2+\sigma_{i,j}^2\right)\cdot \bar{f_{i,j}^2}, \tag{26}$$

where $\bar{f_{i,j}^2} = \mathbb{E}_{\boldsymbol{h}_{i-1}\sim p(\boldsymbol{h}_{i-1})}\left[f_{i,j}(\boldsymbol{h}_{i-1})^2\right]$. Plug this expression back into (24), it follows that

$$\begin{aligned} & \inf_{\boldsymbol{\xi}_i\succ\mathbf{0}} 2\ \mathbb{E}_{\boldsymbol{h}_{i-1}\sim p(\boldsymbol{h}_{i-1})}\left[\mathbb{KL}\left[p(\boldsymbol{h}_i|\boldsymbol{h}_{i-1})||q(\boldsymbol{h}_i)\right]\right] \\ = & \ \mathbb{E}_{\boldsymbol{h}_{i-1}\sim p(\boldsymbol{h}_{i-1})}\left[\sum_j \frac{\left(\mu_{i,j}^2+\sigma_{i,j}^2\right)\cdot f_{i,j}(\boldsymbol{h}_{i-1})^2}{\left(\mu_{i,j}^2+\sigma_{i,j}^2\right)\cdot \bar{f_{i,j}^2}} - \log\frac{\sigma_{i,j}^2\cdot f_{i,j}(\boldsymbol{h}_{i-1})^2}{\left(\mu_{i,j}^2+\sigma_{i,j}^2\right)\cdot \bar{f_{i,j}^2}} - 1\right] \\ = & \ \mathbb{E}_{\boldsymbol{h}_{i-1}\sim p(\boldsymbol{h}_{i-1})}\left[\sum_j \log\frac{\mu_{i,j}^2+\sigma_{i,j}^2}{\sigma_{i,j}^2} + \log\frac{\bar{f_{i,j}^2}}{f_{i,j}(\boldsymbol{h}_{i-1})^2}\right] \\ = & \ \sum_j \log\left(1+\frac{\mu_{i,j}^2}{\sigma_{i,j}^2}\right) + \log\bar{f_{i,j}^2} - \mathbb{E}_{\boldsymbol{h}_{i-1}\sim p(\boldsymbol{h}_{i-1})}\left[\log f_{i,j}(\boldsymbol{h}_{i-1})^2\right] \\ = & \ \sum_j \log\left(1+\frac{\mu_{i,j}^2}{\sigma_{i,j}^2}\right) + \psi_{i,j}. \end{aligned} \tag{27}$$

## Appendix C. Proof of Proposition 1

As a convenient decomposition, we rewrite the objective from (10) using

$$\tilde{\mathcal{L}} = \mathcal{L}_{kl} + \mathcal{L}_{data} \tag{28}$$

$$\mathcal{L}_{kl} \triangleq \sum_{i=1}^{L}\gamma_i\sum_{j=1}^{r_i}\log\left(1+\alpha_{i,j}\right) \tag{29}$$

$$\mathcal{L}_{data} \triangleq -L\int p(\boldsymbol{x},\boldsymbol{y})\log\left(q(\boldsymbol{y}|\boldsymbol{h}_L)\right)\Pi_{i=1}^{L}\Pi_{j=1}^{r_i}p(h_{i,j}|\boldsymbol{h}_{i-1})d\boldsymbol{x}d\boldsymbol{y}d\boldsymbol{h}_{1:L}. \tag{30}$$

We first prove that $\alpha_{i,j}=0$ is a sufficient condition for $\boldsymbol{I}(h_{i,j};\boldsymbol{h}_{i-1})\le\psi_{i,j}$. We have

$$
\begin{aligned}
\boldsymbol{I}(h_{i,j};\boldsymbol{h}_{i-1}) &= \int p(\boldsymbol{h}_{i-1}, h_{i,j}) \log \frac{p(\boldsymbol{h}_{i-1}, h_{i,j})}{p(\boldsymbol{h}_{i-1})p(h_{i,j})} d\boldsymbol{h}_{i-1} dh_{i,j} \\
&\leq \int p(\boldsymbol{h}_{i-1}, h_{i,j}) \log \frac{p(\boldsymbol{h}_{i-1}, h_{i,j})}{p(\boldsymbol{h}_{i-1})q(h_{i,j})} d\boldsymbol{h}_{i-1} dh_{i,j} \\
&= \mathbb{E}_{\boldsymbol{h}_{i-1}\sim p(\boldsymbol{h}_{i-1})}\left[\mathbb{KL}\left[p(h_{i,j}|\boldsymbol{h}_{i-1})||q(h_{i,j})\right]\right], \qquad (31)
\end{aligned}
$$

which holds for any distribution $q$ according to Jensen's inequality. Let

$$
q(h_{i,j}) = \mathcal{N}\left(h_{i,j}|0, \mathbb{E}_{\boldsymbol{h}_{i-1}\sim p(\boldsymbol{h}_{i-1})}\left[f_{i,j}(\boldsymbol{h}_{i-1})^2\right](\mu_{i,j}^2 + \sigma_{i,j}^2)\right). \qquad (32)
$$

Combined with the derivations from Appendix B, we then obtain

$$
\begin{aligned}
\boldsymbol{I}(h_{i,j};\boldsymbol{h}_{i-1}) &\leq \mathbb{E}_{\boldsymbol{h}_{i-1}\sim p(\boldsymbol{h}_{i-1})}\left[\log\left(1 + \frac{\mu_{i,j}^2}{\sigma_{i,j}^2}\right)\right] \\
&\quad + \log\left(\mathbb{E}_{\boldsymbol{h}_{i-1}\sim p(\boldsymbol{h}_{i-1})}\left[f_{i,j}(\boldsymbol{h}_{i-1})^2\right]\right) - \mathbb{E}_{\boldsymbol{h}_{i-1}\sim p(\boldsymbol{h}_{i-1})}\left[\log f_{i,j}(\boldsymbol{h}_{i-1})^2\right] \\
&= \log(1 + \alpha_{i,j}) + \psi_{i,j}. \qquad (33)
\end{aligned}
$$

If $\alpha_{i,j} = 0$, then $\boldsymbol{I}(h_{i,j};\boldsymbol{h}_{i-1}) \leq \psi_{i,j}$.

Now further assume that there exists a ball of radius $\rho > 0$ around a given minimum such that within this ball, $\mathcal{L}_{data}$ is an increasing function of $\sigma_{i,j} \geq 0$. Given this stipulation, if $\alpha_{i,j} = 0$ (which implies that $\mu_{i,j}^2 = 0$), then $\mathcal{L}_{kl}$ will be independent of $\sigma_{i,j}^2$. And so if $\sigma_{i,j}^2 > 0$, there will exist a direction within the ball of radius $\rho$ such that $\mathcal{L}_{data}$ and therefore the overall objective can be reduced, meaning we cannot be at an optimum unless $\sigma_{i,j}^2 = 0$. In this case $\boldsymbol{I}(h_{i,j};\boldsymbol{h}_{i-1}) = 0$ since $p(h_{i,j}|\boldsymbol{h}_{i-1}) = p(h_{i,j})$.

Finally, we prove that for any minimum, $\alpha_{i,j} = 0$ is a necessary condition for $\boldsymbol{I}(h_{i,j};\boldsymbol{h}_{i-1}) = 0$. Let $\{\boldsymbol{W}_i^*, \boldsymbol{b}_i^*, \boldsymbol{\mu}_i^*, \boldsymbol{\sigma}_i^*\}_{i=1}^L$ be a minimum of the objective function, where we have explicitly included a bias term $\boldsymbol{b}_i^*$, and assume $\boldsymbol{I}(\boldsymbol{h}_{i,j};\boldsymbol{h}_{i-1}) = 0$. Then $\boldsymbol{h}_{i,j}$ and $\boldsymbol{h}_{i-1}$ are independent and we have

$$
p(h_{i,j}) = p(h_{i,j}|\boldsymbol{h}_{i-1}) = \mathcal{N}\left(h_{i,j}|f_{i,j}(\boldsymbol{h}_{i-1})\mu_{i,j}, f_{i,j}(\boldsymbol{h}_{i-1})^2\sigma_{i,j}^2\right). \qquad (34)
$$

This suggests that $f_{i,j}(\boldsymbol{h}_{i-1})$ is a constant and $p(h_{i,j})$ is a Gaussian distribution. We denote this constant as $c_{i,j}$ for convenience. We then write the distribution of $p(\boldsymbol{h}_{i+1}|\boldsymbol{h}_{i-1})$ as

$$
\begin{aligned}
&p(\boldsymbol{h}_{i+1}|\boldsymbol{h}_{i-1}) \\
&\quad = \mathbb{E}_{\boldsymbol{h}_i\sim p(\boldsymbol{h}_i|\boldsymbol{h}_{i-1})}\left[\mathcal{N}\left(\boldsymbol{h}_{i+1}|\boldsymbol{\mu}_{i+1}\odot f_{i+1}(\boldsymbol{h}_i), \boldsymbol{\sigma}_{i+1}^2\odot f_{i+1}(\boldsymbol{h}_i)^2\right)\right] \\
&\quad = \mathbb{E}_{\boldsymbol{\epsilon}_i\sim\mathcal{N}(0,I)}\Big[\mathcal{N}\Big(\boldsymbol{h}_{i+1}|\boldsymbol{\mu}_{i+1}\odot f_{i+1}\left(f_i(\boldsymbol{h}_{i-1})\odot(\boldsymbol{\mu}_i + \boldsymbol{\sigma}_i\odot\boldsymbol{\epsilon}_i)\right), \\
&\qquad\qquad \boldsymbol{\sigma}_{i+1}^2\odot f_{i+1}\left(f_i(\boldsymbol{h}_{i-1})\odot(\boldsymbol{\mu}_i + \boldsymbol{\sigma}_i\odot\boldsymbol{\epsilon}_i)\right)^2\Big)\Big] \qquad (35)
\end{aligned}
$$

Note that $f_{i+1}(\cdot)$ is a deterministic function and can be written as

$$
\begin{aligned}
& f_{i+1}(f_i(\boldsymbol{h}_{i-1}) \odot (\boldsymbol{\mu}_i + \boldsymbol{\sigma}_i \odot \boldsymbol{\epsilon}_i)) \\
&\quad = a\left(\sum_{j'} \boldsymbol{W}^*_{i+1,\cdot j'} f_{i,j'}(\boldsymbol{h}_{i-1}) \left(\mu^*_{i,j'} + \sigma^*_{i,j'}\epsilon_{i,j'}\right) + \boldsymbol{b}^*_{i+1}\right) \\
&\quad = a\left(\sum_{j'\neq j} \boldsymbol{W}^*_{i+1,\cdot j'} f_{i,j'}(\boldsymbol{h}_{i-1}) \left(\mu^*_{i,j'} + \sigma^*_{i,j'}\epsilon_{i,j'}\right) + \boldsymbol{W}^*_{i+1,\cdot j} c_{i,j}(\mu^*_{i,j} + \sigma^*_{i,j}\epsilon_{i,j}) + \boldsymbol{b}^*_{i+1}\right), \quad (36)
\end{aligned}
$$

where $a(\cdot)$ is an activation function. We next define a second candidate solution

$$
\begin{aligned}
\boldsymbol{W}^{**}_{i'} &= \boldsymbol{W}^*_{i'}, \\
\mu^{**}_{i',j'} &= \mu^*_{i',j'} \quad \text{if} \quad i' \neq i \text{ or } j' \neq j, \\
\mu^{**}_{i,j} &= 0, \\
\sigma^{**}_{i',j'} &= \sigma^*_{i',j'} \\
\boldsymbol{b}^{**}_{i+1} &= \boldsymbol{b}^*_{i+1} + \boldsymbol{W}^*_{i+1,\cdot j} c_{ij} \mu^*_{i,j}. \qquad (37)
\end{aligned}
$$

At this alternative solution, $f_{i+1}(\cdot)$ becomes

$$
\begin{aligned}
& f_{i+1}(f_i(\boldsymbol{h}_{i-1}) \odot (\boldsymbol{\mu}_i + \boldsymbol{\sigma}_i \odot \boldsymbol{\epsilon}_i)) \\
&\quad = a\left(\sum_{j'} \boldsymbol{W}^{**}_{i+1,\cdot j'} f_{i,j'}(\boldsymbol{h}_{i-1}) \left(\mu^{**}_{i,j'} + \sigma^{**}_{i,j'}\epsilon_{i,j'}\right) + \boldsymbol{b}^{**}_{i+1}\right) \\
&\quad = a\left(\sum_{j'\neq j} \boldsymbol{W}^*_{i+1,\cdot j'} f_{i,j'}(\boldsymbol{h}_{i-1}) \left(\mu^*_{i,j'} + \sigma^*_{i,j'}\epsilon'_{i,j}\right) + \boldsymbol{W}^*_{i+1,\cdot j} c_{i,j}\sigma^*_{i,j}\epsilon_{i,j} + \boldsymbol{W}^*_{i+1,\cdot j} c_{ij}\mu^*_{i,j} + \boldsymbol{b}^*_{i+1}\right). \\
&\qquad (38)
\end{aligned}
$$

This is exactly the same as (36), implying that $p(\boldsymbol{h}_{i+1}|\boldsymbol{h}_{i-1})$ stays the same. Additionally, if we integrate out $\boldsymbol{h}_i$ in $\mathcal{L}_{data}$, we will obtain the same result. Consequently, the data loss at these two solutions is unchanged. And since $\{\boldsymbol{W}^*_i, \boldsymbol{b}^*_i, \boldsymbol{\mu}^*_i, \boldsymbol{\sigma}^*_i\}_{i=1}^L$ is optimal, the value of $\mathcal{L}_{kl}$ must be no greater than at the new solution, *i.e.*, it must be that

$$
\sum_{i'=1} \gamma_{i'} \sum_{j'=1} \log\left(1 + \frac{{\mu^*_{i',j'}}^2}{{\sigma^*_{i',j'}}^2}\right) \leq \sum_{i'=1} \gamma_{i'} \sum_{j'=1} \log\left(1 + \frac{{\mu^{**}_{i',j'}}^2}{{\sigma^{**}_{i',j'}}^2}\right). \qquad (39)
$$

After canceling the equivalent terms, we obtain

$$
\gamma_i \log\left(1 + \frac{{\mu^*_{i,j}}^2}{{\sigma^*_{i,j}}^2}\right) \leq \gamma_i \log\left(1 + \frac{0}{{\sigma^*_{i,j}}^2}\right) = 0. \qquad (40)
$$

So it must be that $\alpha^*_{i,j} = \left(\frac{\mu^*_{i,j}}{\sigma^*_{i,j}}\right)^2 = 0.$ $\square$

## Appendix D. Proof of Proposition 2

If $\boldsymbol{W}_{i+1,\cdot j} = 0$, we have

$$
\begin{aligned}
p(\boldsymbol{h}_{i+1}|\boldsymbol{h}_i) &= \mathcal{N}\left(\boldsymbol{h}_{i+1} \mid \boldsymbol{\mu}_{i+1} \odot \sum_{j'=1}^{r_i} \boldsymbol{W}_{i+1,\cdot j'} h_{i,j'},\;\; \boldsymbol{\sigma}_{i+1}^2 \odot \left(\sum_{j'=1}^{r_i} \boldsymbol{W}_{i+1,\cdot j'} h_{i,j'}\right)^2\right) \\
&= \mathcal{N}\left(\boldsymbol{h}_{i+1} \mid \boldsymbol{\mu}_{i+1} \odot \sum_{j' \neq j} \boldsymbol{W}_{i+1,\cdot j'} h_{i,j'},\;\; \boldsymbol{\sigma}_{i+1}^2 \odot \left(\sum_{j' \neq j} \boldsymbol{W}_{i+1,\cdot j'} h_{i,j'}\right)^2\right) \\
&= p(\boldsymbol{h}_{i+1}|\boldsymbol{h}_{i,\neg j}) \qquad (41)
\end{aligned}
$$

Plug this equation back into (30) we focus on $\Pi_{i'=1}^{L}\Pi_{j'=1}^{r_{i'}} p(h_{i',j'}|\boldsymbol{h}_{i'-1})$ giving

$$
\begin{aligned}
&\Pi_{i'=1}^{L}\Pi_{j'=1}^{r_{i'}} p(h_{i',j'}|\boldsymbol{h}_{i'-1}) \\
&= \left[\Pi_{i'=1}^{i-1} p(\boldsymbol{h}_{i'}|\boldsymbol{h}_{i'-1})\right] \left[\Pi_{j' \neq j} p(h_{i,j'}|\boldsymbol{h}_{i-1})\right] p(h_{i,j}|\boldsymbol{h}_{i-1}) p(\boldsymbol{h}_{i+1}|\boldsymbol{h}_i) \left[\Pi_{i'=i+1}^{L} p(\boldsymbol{h}_{i'}|\boldsymbol{h}_{i'-1})\right] \\
&= \left[\Pi_{i'=1}^{i-1} p(\boldsymbol{h}_{i'}|\boldsymbol{h}_{i'-1})\right] \left[\Pi_{j' \neq j} p(h_{i,j'}|\boldsymbol{h}_{i-1})\right] p(h_{i,j}|\boldsymbol{h}_{i-1}) p(\boldsymbol{h}_{i+1}|\boldsymbol{h}_{i,\neg j}) \left[\Pi_{i'=i+1}^{L} p(\boldsymbol{h}_{i'}|\boldsymbol{h}_{i'-1})\right]. \qquad (42)
\end{aligned}
$$

Note that the only term related to $h_{i,j}$ here is $p(h_{i,j}|\boldsymbol{h}_{i-1})$. So if we plug this back into (30), we can integrate $p(h_{i,j}|\boldsymbol{h}_{i-1})$ out. Hence (30) is independent of $\mu_{i,j}$ and $\sigma_{i,j}$. Again, for any minimum of (28), $\alpha_{i,j}$ should also be a minimum of (29) and thus it should equal to 0. □

## Appendix E. Proof of Proposition 3

First we note that

$$
\begin{aligned}
\int p(\boldsymbol{\epsilon}) \bar{g}(\boldsymbol{\epsilon}; \boldsymbol{\theta})\, d\boldsymbol{\epsilon} &= \int p(\boldsymbol{\epsilon}) \left(\boldsymbol{z}(\boldsymbol{\epsilon};\boldsymbol{\theta})^\top \boldsymbol{A}^\top \boldsymbol{A} \boldsymbol{z}(\boldsymbol{\epsilon};\boldsymbol{\theta}) + \boldsymbol{b}^\top \boldsymbol{z}(\boldsymbol{\epsilon};\boldsymbol{\theta}) + c\right) d\boldsymbol{\epsilon} \qquad (43) \\
&= \int p(\boldsymbol{\epsilon}) \Big[\boldsymbol{\mu}^\top \boldsymbol{A}^\top \boldsymbol{A}\boldsymbol{\mu} + (\boldsymbol{\sigma} \odot \boldsymbol{\epsilon})^\top \boldsymbol{A}^\top \boldsymbol{A} (\boldsymbol{\sigma} \odot \boldsymbol{\epsilon}) + 2 (\boldsymbol{\sigma} \odot \boldsymbol{\epsilon})^\top \boldsymbol{A}^\top \boldsymbol{A} \boldsymbol{\mu} \\
&\qquad + \boldsymbol{b}^\top (\boldsymbol{\mu} + \boldsymbol{\sigma} \odot \boldsymbol{\epsilon}) + c\Big] d\boldsymbol{\epsilon} \\
&\equiv \boldsymbol{\mu}^\top \boldsymbol{A}^\top \boldsymbol{A}\boldsymbol{\mu} + \boldsymbol{b}^\top \boldsymbol{\mu} + \boldsymbol{\sigma}^\top \mathrm{diag}\left[\boldsymbol{A}^\top \boldsymbol{A}\right] \boldsymbol{\sigma} \qquad (44)
\end{aligned}
$$

Hence the upper bound can be effectively simplified to

$$
\bar{\mathcal{L}}(\boldsymbol{\theta}) \equiv \sum_{i=1}^{L} \gamma_i \sum_{j=1}^{r_i} \log\left(1 + \frac{\mu_{i,j}^2}{\sigma_{i,j}^2}\right) + \boldsymbol{\mu}^\top \boldsymbol{A}^\top \boldsymbol{A}\boldsymbol{\mu} + \boldsymbol{b}^\top \boldsymbol{\mu} + \boldsymbol{\sigma}^\top \mathrm{diag}\left[\boldsymbol{A}^\top \boldsymbol{A}\right] \boldsymbol{\sigma}. \qquad (45)
$$

This expression is separable over $\boldsymbol{\sigma}$, and therefore we can optimize with respect to each $\sigma_{i,j}$ individually to compute the resulting penalty function on $\mu_{i,j}$, denoted $\rho(\mu_{i,j}; \gamma_i, a_{i,j})$.

Defining, $a_{i,j}$ to be the diagonal element of $\boldsymbol{A}^\top \boldsymbol{A}$ corresponding with $\sigma_{i,j}$, this gives

$$\begin{aligned}\rho(\mu_{i,j}; \gamma_i, a_{i,j}) &\triangleq \inf_{\sigma_{i,j}>0} \ \gamma_i \log\left(1 + \frac{\mu_{i,j}^2}{\sigma_{i,j}^2}\right) + a_{i,j}\sigma_{i,j}^2 \\ &\equiv \inf_{\sigma_{i,j}>0} \inf_{\xi_{i,j}>0} \ \gamma_i \left(\log \frac{\xi_{i,j}}{\sigma_{i,j}^2} + \frac{\mu_{i,j}^2 + \sigma_{i,j}^2}{\xi_{i,j}}\right) + a_{i,j}\sigma_{i,j}^2 \qquad (46)\end{aligned}$$

after omitting an irrelevant constant. After taking derivatives, equating to zero, and manipulating terms, we find that the unique, optimal value for $\sigma_{i,j}$ is

$$\sigma_{i,j}^* = \left(\frac{a_{i,j}}{\gamma_i} + \frac{1}{\xi_{i,j}}\right)^{-\frac{1}{2}}. \qquad (47)$$

Plugging this value into (46) gives

$$\begin{aligned}\rho(\mu_{i,j}; \gamma_i, a_{i,j}) &\equiv \inf_{\xi_{i,j}>0} \ \gamma_i \left[\log \xi_{i,j} + \log\left(\frac{a_{i,j}}{\gamma_i} + \frac{1}{\xi_{i,j}}\right) + \frac{\mu_{i,j}^2}{\xi_{i,j}}\right] \\ &\equiv \inf_{\xi_{i,j}>0} \ \gamma_i \left[\log\left(\frac{\gamma_i}{a_{i,j}} + \xi_{i,j}\right) + \frac{\mu_{i,j}^2}{\xi_{i,j}}\right] \qquad (48)\end{aligned}$$

again excluding constants. Moreover, it has been shown in Wipf et al. (2011) that the function defined as

$$f_\delta(x) \triangleq \inf_{\alpha>0} \frac{x^2}{\alpha} + \log(\alpha + \delta) \qquad (49)$$

will be concave and non-decreasing in $|x|$ for all $\delta \geq 0$, and therefore, $\rho(\mu_{i,j}; \gamma_i, a_{i,j})$ must be concave and non-decreasing with respect to $|\mu_{i,j}|$ since $\gamma_i / a_{i,j}$ is non-negative.

Proceeding further, if any $\boldsymbol{\theta}^* = \{\boldsymbol{\mu}^*, \boldsymbol{\sigma}^*\}$ is any minimizer of $\bar{\mathcal{L}}(\boldsymbol{\theta})$ (local or global), it follows from the developments above that $\boldsymbol{\mu}^*$ must be at least a local minimum of

$$\inf_{\boldsymbol{\sigma}\succ\mathbf{0}} \bar{\mathcal{L}}(\boldsymbol{\theta}) \ \equiv \ \sum_{i=1}^{L}\sum_{j=1}^{r_i} \rho(\mu_{i,j}; \gamma_i, a_{i,j}) + \boldsymbol{\mu}^\top \boldsymbol{A}^\top \boldsymbol{A}\boldsymbol{\mu} + \boldsymbol{b}^\top \boldsymbol{\mu}. \qquad (50)$$

Now let

$$\boldsymbol{u} \triangleq \boldsymbol{A}\boldsymbol{\mu}^*, \quad v \triangleq \boldsymbol{b}^\top \boldsymbol{\mu}^*, \quad \widetilde{\boldsymbol{A}} \triangleq \begin{bmatrix} \boldsymbol{A} \\ \boldsymbol{b}^\top \end{bmatrix}, \ \text{ and } \ \widetilde{\boldsymbol{u}} \triangleq \begin{bmatrix} \boldsymbol{u} \\ v \end{bmatrix} \qquad (51)$$

for any arbitrary local minimum. It automatically follows that $\boldsymbol{\mu}^*$ is at least a local feasible solution to

$$\inf_{\boldsymbol{\mu}} \ \sum_{i=1}^{L}\sum_{j=1}^{r_i} \rho(\mu_{i,j}; \gamma_i, a_{i,j}) \quad \text{s.t.} \quad \widetilde{\boldsymbol{u}} = \widetilde{\boldsymbol{A}}\boldsymbol{\mu}. \qquad (52)$$

If it were not, then we could adjust $\boldsymbol{\mu}^*$ along a feasible direction to reduce (52), which would necessarily further reduce $\bar{\mathcal{L}}(\boldsymbol{\theta})$ leading to a contradiction. Furthermore, solving (52) amounts to minimizing a separable objective function, concave and non-decreasing in the magnitudes of the elements in $\boldsymbol{\mu}$ across a set of linear constraints. As shown in Rao et al. (2003), the local minima of problems in this form will have at most rank$[\widetilde{\boldsymbol{A}}]$ nonzero elements . This implies that

$$\|\boldsymbol{\mu}^*\|_0 \leq \text{rank}[\widetilde{\boldsymbol{A}}] \leq \text{rank}[\boldsymbol{A}] + 1. \tag{53}$$

We have not though as of yet demonstrated that

$$\|\boldsymbol{\mu}^*\|_0 = \|(\boldsymbol{\mu}^*)^2 \odot (\boldsymbol{\sigma}^*)^{-2}\|_0. \tag{54}$$

For example, if $\boldsymbol{\sigma}$ likewise has elements converging to zero for certain indices $i$ and $j$, then $\mu_{i,j}^2 \sigma_{i,j}^{-2}$ is indeterminate. We must therefore consider limiting cases to establish (54). For this purpose, note that any locally minimizing $\sigma_{i,j}^*$ must satisfy (47). From this expression we may conclude that if some locally-minimizing $\mu_{i,j}^* = 0$ is obtained while the corresponding $\xi_{i,j}^* > 0$, then likewise $\sigma_{i,j}^* > 0$ and we may safely conclude that $\mu_{i,j}^2 \sigma_{i,j}^{-2} = 0$. So we need only further consider the case where $\xi_{i,j}^* \to 0$, a necessary condition for $\sigma_{i,j}^* \to 0$ at any local minimum based on (47).

Based on earlier results above, when conditioned on $\boldsymbol{\xi}$ any locally minimzing $\boldsymbol{\mu}$ must satisfy

$$\begin{aligned}
\boldsymbol{\mu}^* &= \inf_{\boldsymbol{\mu}} \; \boldsymbol{\mu}^\top \boldsymbol{A}^\top \boldsymbol{A} \boldsymbol{\mu} + \boldsymbol{b}^\top \boldsymbol{\mu} + \gamma_i \sum_{i=1}^{L} \boldsymbol{\mu}_i^\top \text{diag}[\boldsymbol{\xi}_i]^{-1} \boldsymbol{\mu}_i \\
&= \inf_{\boldsymbol{\mu}} \; \boldsymbol{\mu}^\top \left( \boldsymbol{A}^\top \boldsymbol{A} + \boldsymbol{D}^{-1} \right) \boldsymbol{\mu} + \boldsymbol{b}^\top \boldsymbol{\mu} \\
&= \left( \boldsymbol{A}^\top \boldsymbol{A} + \boldsymbol{D}^{-1} \right)^{-1} \boldsymbol{b} \\
&= \boldsymbol{D} \left( \boldsymbol{A}^\top \boldsymbol{A} \boldsymbol{D} + \boldsymbol{I} \right)^{-1} \boldsymbol{b},
\end{aligned} \tag{55}$$

where $\boldsymbol{D}$ is a diagonal matrix defined such that the first and second lines are equivalent. Based on the right multiplication by $\boldsymbol{D}$ on the final line, it follows that $\mu_{i,j}^2 = O\left(\xi_{i,j}^2\right)$, i.e., the value cannot be larger than $\xi_{i,j}^2$ up to a constant. Combining with (47), this implies that $\mu_{i,j}^2 \sigma_{i,j}^{-2} = O(\xi_{i,j}) \to 0$ when $\xi_{i,j} \to 0$. Therefore it must be that $\|\boldsymbol{\mu}^*\|_0 \geq \|(\boldsymbol{\mu}^*)^2 \odot (\boldsymbol{\sigma}^*)^{-2}\|_0$. As for the other direction, because it is easily demonstrated that $\boldsymbol{\xi}$ must be bounded at any local minimum, we cannot have any $\sigma_{i,j} \to \infty$, and therefore if $\mu_{i,j}^2 \sigma_{i,j}^{-2} = 0$, it must be that $\mu_{i,j} = 0$ satisfying the final requirement of (54). □